\documentclass{article}

\usepackage{arxiv}

\usepackage[T1]{fontenc}
\usepackage[utf8]{inputenc}
\usepackage{graphicx}
\usepackage{booktabs}
\usepackage{amsmath,amssymb}
\usepackage{cite}          % sorts + compresses multiple citations
\usepackage[hidelinks]{hyperref}
\usepackage{orcidlink}     % \orcidlink for author ORCID iDs
\usepackage{xcolor}
\usepackage{microtype}

\graphicspath{{figures/}}

\begin{document}

\title{Physics-Informed Machine Learning Under Small-Data Constraints: Lessons from Abrasive Waterjet Milling%
\thanks{Accepted for publication; to be published in the Proceedings of ICANN 2026, Springer LNCS.}}

% Short title for the running header (full title is too long for the header)
\renewcommand{\shorttitle}{Physics-Informed ML Under Small-Data Constraints}

\author{%
  Sarah Grewe\,\orcidlink{0009-0009-7572-826X} \\
  Bochum University of Applied Sciences\\
  Germany\\
  \texttt{sarah.grewe@hs-bochum.de} \\
  \And
  Jörg Frochte\,\orcidlink{0000-0002-5908-5649} \\
  Bochum University of Applied Sciences\\
  Germany\\
  \texttt{joerg.frochte@hs-bochum.de} \\
}

\date{}

\maketitle

\begin{abstract}

In physically dominated machining processes, experimental datasets are
small, expensive, and material-specific; in this regime, data
curation, evaluation design, and the form of physics integration can
matter as much as the learning algorithm. 
Using an abrasive waterjet milling dataset ($n{=}155$, Inconel\,718),
we make three methodological contributions.  First, we separate
physics-based data \emph{cleaning} from statistical \emph{curation}
and treat the latter as competing modelling hypotheses rather than
silent preprocessing.  Second, we find that model rankings from a
15-point hold-out set can be unstable: the single-split winner drops
from rank~1 to rank~7 under 10-fold cross-validation, while Gaussian
Process (GP) variants occupy the top ranks.  Third, we study a
spectrum of physics integration levels and find that residual
learning on a compact physics baseline is competitive for GP,
yielding lower variance and an interpretable decomposition, but
degrades tree-based models. 
Bayesian hyperparameter tuning improves parameter-sensitive baselines
such as gradient boosting and SVR, yet harms multi-stage hybrid
pipelines at this sample size.  GP uncertainty intervals are
approximately calibrated ($86\%$ empirical coverage at nominal
$90\%$).  The resulting picture is methodological: for small, expensive process datasets, our results suggest that, in this setting, reliable model comparison benefits from explicit curation hypotheses, robust evaluation, and careful choices about how physics enters the model.

\end{abstract}


\section{Introduction}
\label{sec:intro}

When a high-pressure water jet loaded with abrasive particles removes material in a controlled fashion--- known as abrasive waterjet milling (AWJM)---the goal is not to sever the workpiece but to achieve a prescribed \emph{milling depth}~$\bar d$ with tight tolerances. Three process parameters dominate the outcome: water pressure~$p$, traverse speed~$v$, and abrasive mass flow rate~$\dot m$~\cite{Momber1997,Alberdi2010}.  Finding the right combination for a target depth currently requires expensive trial runs -- a task for which data-driven surrogate models have been proposed. However, AWJM datasets are inherently small -- typical studies report 27--50 measurements~\cite{Prabhu2021,prabhu2024,Chen2024} -- since each material and machine configuration requires its own experiments, and a single data point can cost hours of machine time plus post-process profilometry.
Our dataset contains $n{=}155$ measurements on Inconel\,718, a nickel-based superalloy common in aerospace manufacturing. This limitation is structural, not temporary: transferring models to a new material requires fresh experimentation, and the cost per data point remains considerable.
Small datasets create a cascade of methodological challenges that are rarely examined:  First, \emph{data quality}:  with 155 points, a handful of recording errors or regime transitions can shift model rankings, yet preprocessing choices are seldom treated as modelling decisions~\cite{Zhao2024}. Second,  \emph{evaluation reliability}: a held-out set of 10--15 points is too small for stable model comparison~\cite{Raschka2018,Varoquaux2018}, making single-split evaluation insufficient for stable model comparison in this setting. Third, \emph{physics integration}: semi-empirical AWJM models~\cite{Hashish1989,Hlavac2021} capture dominant trends but require material-specific calibration; whether they improve ML predictions remains open.
We argue that in this small-data regime, modelling decisions begin \emph{before} model fitting: choice of anomaly identification, statistical filtering, and evaluation design can change conclusions as much as algorithm choice itself.

\paragraph{Contributions.}
Using the Inconel\,718 AWJM dataset, we make three methodological contributions. First, we separate physics-based data \emph{cleaning} (removing demonstrably erroneous points) from statistical \emph{curation} (filtering extreme but plausible points) and treat curation as competing modelling hypotheses. Second, we find that the evaluation protocol is itself a modelling choice: a 15-point hold-out set produces a different ranking than 10-fold cross-validation, with the single-split winner dropping to rank~7. Third, we study four physics integration levels and find that residual physics helps smooth interpolators (GP) but degrades tree-based methods; Bayesian hyperparameter tuning helps parameter-sensitive baselines yet harms multi-stage hybrids,
and GP intervals remain only approximately calibrated. The contribution is methodological: a controlled analysis of how these upstream choices interact in the small-data
regime.

\section{Related Work}
\label{sec:related}

\paragraph{AWJM surrogate modelling.}
ML-based surrogate modelling for AWJM has been explored with neural networks~\cite{Deng2023}, ensemble methods~\cite{prabhu2024,Chen2024}, and support vector regression~\cite{Prabhu2021}; yet probabilistic approaches such as Gaussian process regression---despite demonstrated potential for physics-informed modelling in adjacent domains~\cite{Cross2024}---have received little attention. Zhao et al.~\cite{Zhao2024} identify data quality as an under-addressed challenge in ML-based manufacturing, consistent with the present study. 

\paragraph{Small-data ML in manufacturing.}
The small-data regime ($n < 200$) is the norm in materials science and process engineering. Xu et
al.~\cite{Xu2023}  and Fullington et al. ~\cite{Fullington2024} survey strategies such as transfer learning, data augmentation, and physics priors for small-data ML in materials science and additive manufacturing, respectively; these help but do not eliminate the fundamental limitation. For tabular regression problems like AWJM depth prediction, classical methods such as Gaussian Processes remain competitive~\cite{Grinsztajn2022,ShwartzZiv2022}. 
Cross-validation instability at small $n$ is well documented: Vabalas et al~\cite{Vabalas2019} show that $k$-fold CV produces biased performance estimates even at $n{=}1000$, with bias worsening when feature selection is performed on pooled data.

\paragraph{Physics-informed and hybrid modelling.}
In this paper, we use the term physics-informed machine learning in the broader sense of scientific-knowledge integration into the learning pipeline, including feature representations, residual formulations, and prior mean functions, rather than restricting it to physics-informed loss terms in neural networks.
Physics can enter ML pipelines at various stages: Willard et al.~\cite{Willard2022} provide a taxonomy distinguishing physics-guided loss functions, initialisation, architecture design, and hybrid models, while Cross et al.~\cite{Cross2024} formalise a spectrum of physics-informed Gaussian Processes from physics-derived kernels and mean functions to constrained covariance formulations.
For AWJM, semi-empirical models date back to Hashish~\cite{Hashish1984,Hashish1989}, later revised by Hlav'{a}\v{c}~\cite{Hlavac2021}; these capture dominant trends but still require material-specific calibration. The open question is not whether physics \emph{can} be integrated, but under which conditions -- algorithm class, integration level, data curation -- it yields measurable improvement.

\paragraph{Evaluation and tuning at small~$n$.}
Cawley and Talbot~\cite{Cawley2010} show that overfitting in model selection can cause performance degradation comparable inter-algorithm differences---a risk amplified at small~$n$. The distinction between aleatory and epistemic uncertainty~\cite{DerKiureghian2009} is particularly relevant here: epistemic uncertainty dominates, and calibrating prediction intervals requires more held-out data than is typically available~\cite{Angelopoulos2023}.

\section{Dataset and Data Preparation}
\label{sec:data}

\subsection{Dataset}

The dataset comprises $n{=}155$ single-pass AWJM experiments on Inconel\,718 with five pressure levels
($p \in \{1000,\ldots,5000\}$\,bar), six traverse speeds ($v \in \{250,\ldots,3000\}$\,mm/min), and six abrasive mass flow rates ($\dot m \in \{60,\ldots,480\}$\,g/min).  The target is the mean milling depth~$\bar d$ (mm), ranging from
$-5.5$\,mm to near zero. The design is not fully crossed: of the $5\times6\times6=180$ nominal parameter combinations, 155 were realised. The distribution of $\bar d$ is left-skewed (skewness $-1.28$).

\subsection{Two-Stage Data Preparation}
\label{sec:twostage}

We distinguish two operations commonly conflated in applied ML studies.
\emph{Stage~1 (physics-based cleaning).}
Domain knowledge identifies 25 demonstrably erroneous or unreliable points: 
(a)~13 with physically impossible positive depth (sign anomalies from a recording error);
(b)~9 with within-experiment variance exceeding the mean depth (variation coefficient $> 1$; one overlaps with~(a));
(c)~4 at a regime boundary where cutting degrades into surface glazing. 
These are always excluded.
\emph{Stage~2 (statistical curation as competing hypotheses).}
The remaining 130 points are physically plausible but statistically extreme (e.g.\ very deep cuts at low traverse speed and high pressure). Whether excluding them improves prediction depends on the algorithm's bias--variance trade-off. Rather than committing to a single filter, we define four hypotheses (Table~\ref{tab:hypotheses}) and
evaluate each by its downstream effect on model performance.
\begin{table}[h!t]
\caption{Competing outlier hypotheses applied per pressure group
  (Stage-1 flags always retained; $n_\text{train}$ as defined in Section~\ref{sec:eval}).
  MAD: median absolute deviation~\cite{Hampel1974}; a point is
  flagged if its modified $Z$-score
  $M_i = 0.6745\,(y_i - \tilde y)/\text{MAD}$ exceeds the
  threshold~$\tau$.
  IQR: interquartile range; points outside the Tukey fence
  $[Q_1 - 1.5\,\text{IQR},\; Q_3 + 1.5\,\text{IQR}]$ are removed.}
\label{tab:hypotheses}
\centering\small
\begin{tabular}{llr}
\toprule
Hypothesis & Method & $n_\text{train}$ \\
\midrule
H-NONE    & No statistical filter & 115 \\
H-MAD-3.5 & MAD, $\tau{=}3.5$ & 99 \\
H-MAD-2.5 & MAD, $\tau{=}2.5$ & 97 \\
H-IQR     & Tukey fence, $1.5\times$IQR & 105 \\
\bottomrule
\end{tabular}
\end{table}
The modified $Z$-score assumes approximate normality; since filtering is applied per pressure group ($n \approx 30$),
within-group distributions are considerably less skewed than the overall dataset. The thresholds $\tau{=}2.5$~\cite{Leys2013} and $\tau{=}3.5$~\cite{Iglewicz1993} bracket a moderately conservative to permissive range. 
Statistical filtering is justified not by the method itself but by its downstream effect on prediction quality.

\emph{Target transformation.}
The signed square root $y_t = \sqrt{|y|}\,\text{sgn}(y)$
reduces skewness from $-1.28$ to $-0.10$ while preserving sign, remaining parameter-free and invertible. It provided a consistent 5--9\% RMSE reduction across all algorithms; all results below use the transformed target.

\section{Methodology}
\label{sec:method}

\subsection{Physics Baseline Model}
\label{sec:physics}
Milling depth approximately follows a power law in the process parameters. 
 Hashish~\cite{Hashish1984,Hashish1989} established that jet kinetic energy scales as $p^{3/2}$ (Bernoulli), depth increases with abrasive loading~$\dot m^\gamma$, and is inversely proportional to traverse speed~$v$.
However, a simple power law $\bar d = -C\, p^\alpha \dot m^\gamma / v$ overestimates depth at high pressures because the cutting efficiency saturates~\cite{Momber1997,Hlavac2021}.  Following Hlav\'{a}\v{c}'s observation that pressure-dependent saturation must be modelled explicitly~\cite{Hlavac2021}, we introduce a compact rational denominator:
\begin{equation}\label{eq:pgh}
  \bar d \;=\; -\,\frac{C \;\cdot\; p^{3/2} \;\cdot\;
    \dot m^{\,\gamma}}{v \;\cdot\; (A\,p + B)}\,,
\end{equation}
where $C$, $A$, $B$, $\gamma$ are fitted via nonlinear least squares (soft-$L_1$ loss, 10 random restarts).
The pressure exponent $\alpha = 3/2$ is fixed from Bernoulli's equation; $\gamma$ is fitted because abrasive efficiency is material-dependent. The denominator $(Ap + B)$ introduces pressure-dependent resistance: at low $p$, $B$ dominates (constant resistance); at high $p$, $Ap$ dominates and depth growth decelerates. We refer to this 4-parameter model as PG-H(4) (physics-guided, Hlaváč-inspired, 4 parameters).
PG-H(4) serves as a \emph{physics backbone} for hybrid models, not as a standalone predictor. With four free parameters, it captures roughly half the variance ($R^2 \approx 0.51$ on training data), encoding the dominant scaling so that the ML residual $r = y - f_\text{phys}(x)$ is simpler and lower-variance than the raw target.

\subsection{Physics Integration Levels}
\label{sec:levels}

Rather than treating physics integration as binary, we define four levels differing in how much prior
knowledge the ML model receives: At \emph{Level~0} (pure ML): raw features ($p$, $v$, $\dot m$); model must learn the full input--output mapping; distance-based methods (GP, SVR) use standardised features, tree-based methods receive raw features (splits are invariant to monotone rescaling). At \emph{Level~1} (log-transform): features $(\ln p, \ln v, \ln \dot m)$, encoding the power-law assumption so the  pure power law becomes linear. For tree-based methods the monotone transform produces identical splits, making Levels~0 and~1 equivalent; for kernel methods, it changes the distance metric. At \emph{Level~2} (residual learning): PG-H(4) provides a first approximation~$f_\text{phys}(x)$ and the ML model learns the additive residual $r = y - f_\text{phys}(x)$, yielding $\hat y = f_\text{phys}(x) + \hat r(x)$. The physics model need not be accurate in every detail -- it suffices that it captures dominant trends. At \emph{Level~3} (PG-H(4) as GP prior mean): in data-rich regions the posterior corrects the physics; in data-sparse regions the prediction reverts to the physics model---the strongest integration but most sensitive to physics model accuracy.
The requirements on the physics model differ across levels. At Levels~0--2, the ML explicitly corrects physics error; inaccurate assumptions are tolerable. At Level~3, reversion to the physics in data-sparse regions means accuracy directly governs extrapolation quality.

\subsection{ML Algorithms}
\label{sec:algorithms}

Given small~$n$, we prioritise sample efficiency over model capacity.  We evaluate 8~algorithms across all integration levels; the main comparison focuses on:
\emph{Gaussian Processes (GP)} with a Mat\'{e}rn-5/2 kernel and automatic relevance determination; hyperparameters are optimised by marginal likelihood maximisation~\cite{Rasmussen2006}, with $\mathcal{O}(n^3)$ cost negligible at $n \approx 100$. GP natively provides predictive variance.
\emph{Gradient Boosting (GB)}, configured conservatively
(\texttt{max\_depth}$=3$, 100~iterations, learning rate~$0.05$,
subsample~$0.8$) to limit overfitting.
\emph{SVR} with RBF kernel ($\varepsilon{=}0.05$, $C \in \{0.1, 1, 10, 100\}$, 5-fold grid search).
\emph{Ridge regression} as linear baseline ($\alpha \in \{0.01, 0.1, 1, 10, 100\}$, grid search). 
Four additional tree-based methods (XGBoost, LightGBM, HistGBR, Random Forest) ranked below all GP and standard GB variants under 10-fold cross-validation and are reported  only in aggregate. 
A shallow MLP baseline (two hidden layers, $3{\to}32{\to}16{\to}1$, ${\sim}625$ parameters) with MC-Dropout~\cite{Gal2016} was included as an uncertainty-quantification comparison only, not as a methodological focus; its best configuration (Level~2, H-NONE) achieved test RMSE${=}0.55$\,mm, ranking below all GP variants but providing an approximate uncertainty estimate for comparison (Sec.~\ref{sec:uncertainty}).
Non-grid-search models use fixed hyperparameters; at $n \approx 100$, 5-fold CV for tuning leaves
only ${\sim}16$ points per validation fold -- too few for a reliable signal (Sec.~\ref{sec:tuning}).

\subsection{Evaluation Protocol}
\label{sec:eval}
After Stage-1 cleaning ($155 \to 130$ points), 15~points are reserved as a fixed, stratified test set, leaving $115$ for training. Of these, $111$ survive \emph{every} Stage-2 filter, forming the ``always clean'' pool ($96$ training, $15$ test). Stage-2 curation is applied only within the
training partition: filters are computed from training data alone, preventing information leakage.
In the \emph{single-split} setting, the 15~held-out points serve as
test set. In the \emph{10-fold CV} setting, the 96 always-clean training points are partitioned into 10~folds; the remaining 19 points (removed by at least one Stage-2 filter) augment each fold's training set when they pass the active filter. Stage-2 filters are recomputed per fold from training data only. We report RMSE~(mm) for overall accuracy and MAE~(mm) as a scale-independent complement less sensitive to outliers.

\section{Results}
\label{sec:results}
\subsection{Single-Split vs.\ Multi-Fold Rankings}
\label{sec:singlevsmulti}
Table~\ref{tab:ranking} contrasts single-split and 10-fold results
for the top configurations.  The single-split champion (GB, H-IQR,
RMSE $0.262$) drops to rank~7 under 10-fold CV
($0.520 \pm 0.220$), while GP variants occupy the top~7 ranks.
\begin{table}[h!t]
\caption{Rank reversal between a 15-point hold-out set and 10-fold CV, sorted by 10-fold CV
RMSE; L0: pure ML (Level~0), L2: residual learning on PG-H(4) (Level~2); see Section~\ref{sec:levels}}
\label{tab:ranking}
\centering\small
\begin{tabular}{llrrrrrr}
\toprule
Model & Hyp.
  & Rank$_\text{1-split}$
  & Rank$_\text{10-fold}$
  & \multicolumn{1}{c}{RMSE$_\text{1-split}$}
  & \multicolumn{1}{c}{RMSE$_\text{10-fold}$}
  & \multicolumn{1}{c}{$\pm$std}
  & \multicolumn{1}{c}{MAE} \\
\midrule
GP L0        & MAD-3.5 & 3 & \textbf{1} & 0.324 & \textbf{0.411} & 0.232 & 0.261 \\
GP L2+PGH4   & MAD-2.5 & 6 & 2 & 0.467 & 0.432 & \textbf{0.150} & 0.267 \\
GP L0        & MAD-2.5 & 5 & 3 & 0.374 & 0.437 & 0.181 & 0.311 \\
GP L0        & IQR     & 4 & 4 & 0.343 & 0.446 & 0.244 & 0.295 \\
GB L0        & IQR     & \textbf{1} & 7 & \textbf{0.262} & 0.520 & 0.220 & 0.344 \\
\bottomrule
\end{tabular}
\end{table}
This is not a marginal reshuffling but a substantive rank reversal. At $n \approx 100$, a 15-point held-out set yields point estimates with apparent precision, yet the fold variance suggests that, in this setting, these point estimates should be interpreted cautiously and may not support stable model comparison.
\subsection{Effect of Data Curation}
The outlier hypothesis interacts with algorithm class: GP, as a smooth interpolator, prefers cleaner data (H-MAD-3.5 or H-MAD-2.5), accepting fewer points for lower noise; tree-based methods prefer more data (H-NONE or H-IQR), being inherently robust to outliers. The best-to-worst hypothesis gap exceeds $0.10$\,mm RMSE for GP -- to inter-algorithm differences. The sqrt target transformation provides a consistent but modest benefit ($5$--$9\%$ RMSE reduction).
\subsection{Physics Integration}
\label{sec:physics_results}
\begin{figure}[h!t]
  \centering
  \vspace{-7mm}
  \includegraphics[width=\textwidth]{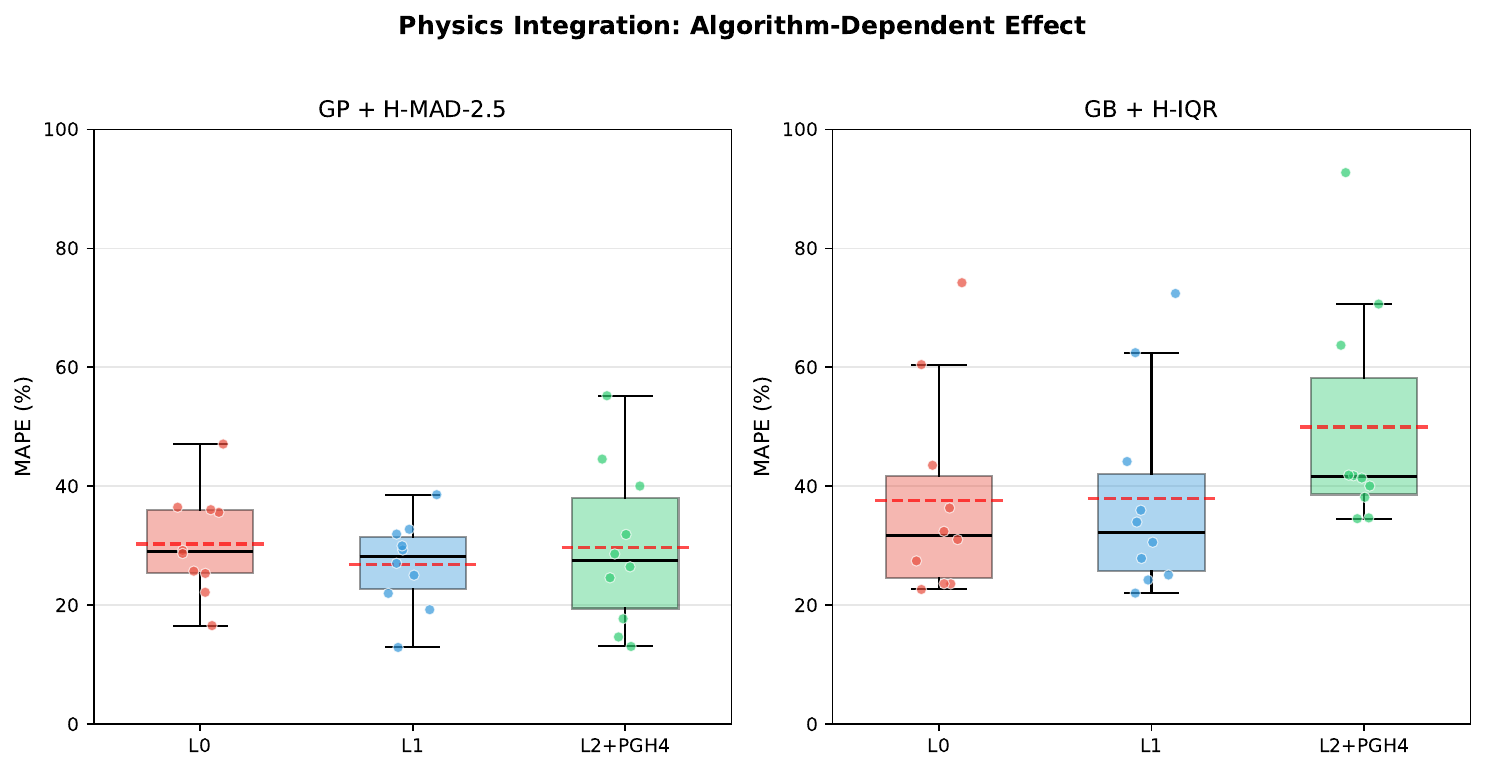}
  \caption{Physics integration effect on MAPE across 10~folds.
    For GP, Level~2 is competitive (similar median, lower variance);
    for GB, Level~2 increases both median and spread.
    MAPE is included as it exposes GB degradation more clearly than RMSE, which is dominated by deep cuts.}
  \label{fig:physics}
\end{figure}
Physics integration is algorithm-dependent (Fig.~\ref{fig:physics}). For GP with H-MAD-2.5, Level~2 (residual learning) achieves RMSE $0.432 \pm 0.150$ vs. $0.513 \pm 0.233$ for pure GP at Level~0 (MAPE $30\% \pm 9\%$ vs.\ $30\% \pm 14\%$). A paired $t$-test detects no significant mean difference ($p = 0.14$), but the hybrid's fold-to-fold variance is notably lower (Levene $p = 0.06$, variance ratio $2.4{:}1$); with only 10~folds neither test has high power, so these results are indicative rather than conclusive. For GB, Level~2 increases RMSE from $0.524 \pm 0.225$ to $0.549 \pm 0.216$ and MAPE from $38\% \pm 17\%$ to $50\% \pm 19\%$; the physics residuals lack the axis-aligned structure that trees exploit. MAPE is shown because relative error exposes GB degradation more clearly than RMSE, which is dominated by deep cuts.
The hybrid GP decomposes each prediction as $\hat y = f_\text{phys}(x) + \hat r(x)$.  At shallow cuts
($|\bar d| < 0.3$\,mm), the physics component accounts for ${\sim}81\%$ of the prediction; at deep cuts ($>1$\,mm), the learned residual contributes ${\sim}40\%$. This decomposition quantifies how much each prediction relies on the calibrated physics model versus the data-driven correction term.
A training-size ablation (subsampling at 20\%, 40\%, 60\%, 80\%, 100\%) does not confirm that physics integration provides a larger benefit when data are scarce: at 20--60\% training size, L0 and L2 fold-to-fold standard deviations are comparable (ratio ${\leq}\,1.2{:}1$). The variance advantage of the hybrid emerges only at full data size ($1.6{:}1$).
Prediction accuracy is strongly depth-dependent. For deep cuts ($|\bar d| > 1$\,mm) the best GP achieves
MAE~$\approx 0.18$\,mm, whereas shallow cuts ($|\bar d| < 0.3$\,mm) reach MAE~$\approx 0.05$\,mm ---small in absolute terms but a large fraction of the target. This is an inherent signal-to-noise limitation: at $|\bar d| \approx 0.1$\,mm, even $0.02$\,mm absolute error -- well below the process standard deviation -- represents $20\%$ of the target.
\subsection{Hyperparameter Tuning at Small~$n$}
\label{sec:tuning}
Bayesian optimisation (Optuna~\cite{optuna2019}, 60 trials,
5-fold inner CV) was applied to six configurations.  Results
(Table~\ref{tab:optuna}) show that: 
\begin{table}[h!t]
\caption{Effect of Optuna hyperparameter tuning (nested CV).}
\label{tab:optuna}
\centering\small
\begin{tabular}{lrrr}
\toprule
Model & Default RMSE & Tuned RMSE & $\Delta$ \\
\midrule
GP L0 (MAD-3.5)  & 0.411 & 0.414 & $+0.003$ \\
GP L2+PGH4 (MAD-2.5)       & 0.432 & 0.504 & $+0.072$ \\
GB L0 (IQR)      & 0.524 & 0.431 & $-0.093$ \\
SVR L0 (MAD-3.5) & 0.534 & 0.456 & $-0.078$ \\
\bottomrule
\end{tabular}
\end{table}
GP is unaffected because kernel hyperparameters are already optimised via marginal likelihood. Algorithms with sensitive defaults (GB, SVR) improve substantially ($-0.078$ to $-0.093$\,mm RMSE).  The hybrid GP~L2, however, degrades: the two-stage pipeline (physics fit $\to$ residual GP) introduces a search space where ${\sim}16$ validation points per fold are insufficient for reliable selection. For small dataset settings, automated tuning should thus be applied selectively: it benefits models with sensitive hyperparameters but can degrade multi-stage pipelines whose inner validation sets are too small for reliable selection.
\subsection{Uncertainty Quantification}
\label{sec:uncertainty}
GP models provide native predictive variance.  Because we model the sqrt-transformed target, uncertainty is back-propagated via the delta method: $\sigma_y \approx 2|y_t|\,\sigma_{y_t}$.  This first-order approximation underestimates variance when $|y_t| \to 0$, partially explaining degraded coverage at small depths.  Calibration is assessed across 10~folds (Table~\ref{tab:calibration}).
\begin{table}[h!t]
\caption{Empirical coverage (\%) at nominal confidence levels.}
\label{tab:calibration}
\centering\small
\begin{tabular}{lcccccc}
\toprule
 & 50\% & 60\% & 70\% & 80\% & 90\% & 95\% \\
\midrule
GP L0 (standalone) & 47 & 56 & 65 & 75 & 86 & 92 \\
GP L2+PGH4 (hybrid) & 37 & 45 & 55 & 65 & 74 & 82 \\
MLP MC-Dropout & 40 & 47 & 47 & 53 & 67 & 73 \\   
\bottomrule
\end{tabular}
\end{table}
GP~L0 achieves $86\%$ empirical coverage at nominal $90\%$ (interval width $1.04$\,mm), the hybrid GP~L2 drops to $74\%$, and MC-Dropout on the MLP (100 stochastic forward passes) falls to $67\%$ -- confirming that, at this sample size, analytically derived GP variance outperforms both hybrid and approximate neural-network uncertainty. Formally, $\sigma^2_\text{hybrid}$ captures only residual uncertainty $\text{Var}[\hat r(x)]$ but omits physics model-form uncertainty $\text{Var}[f_\text{phys}(x)]$~\cite{DerKiureghian2009}; incorporating the latter requires a probabilistic physics baseline (future work). 
\begin{figure}[h!t]
  \centering
  \vspace{-3mm}
  \includegraphics[width=\textwidth]{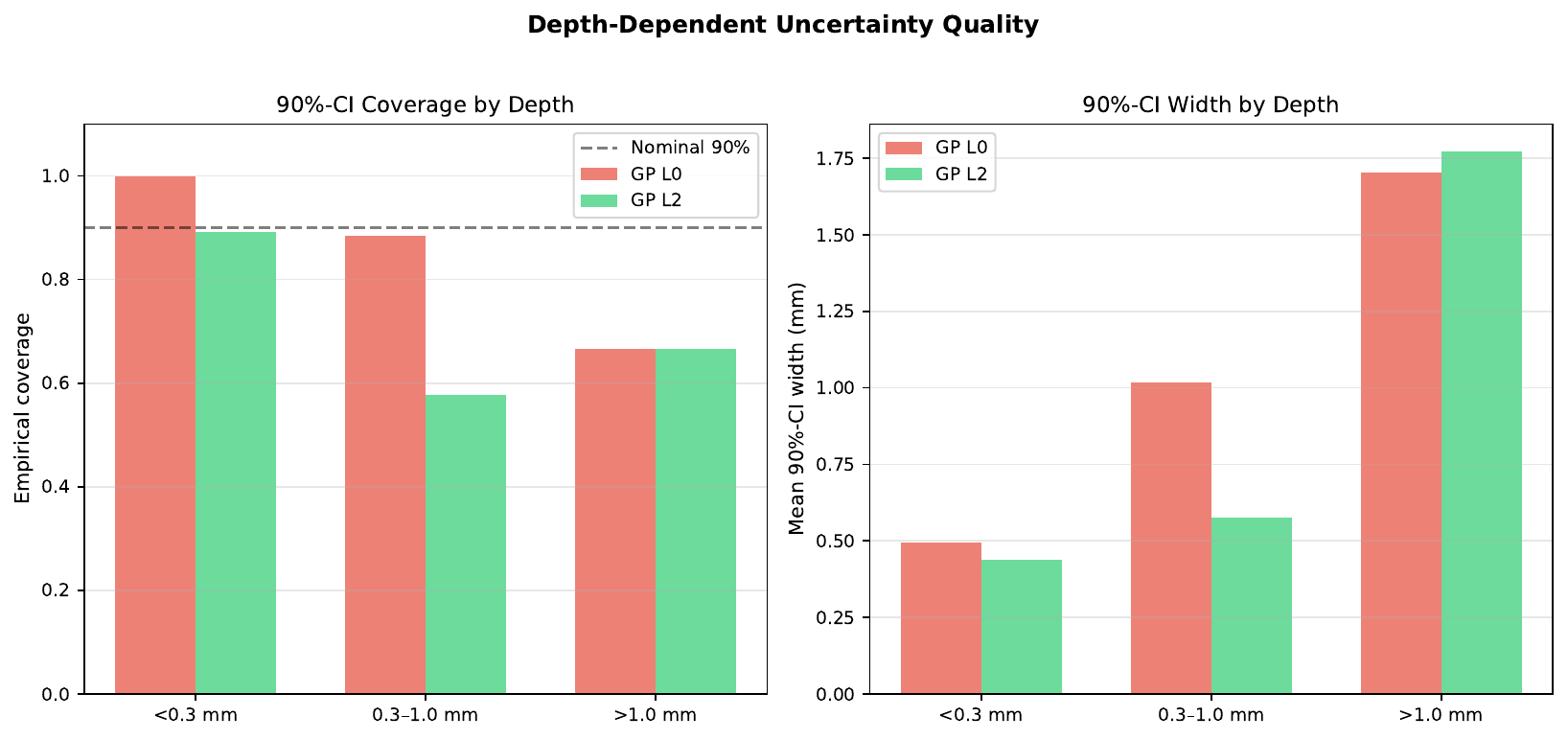}
  \caption{Depth-dependent uncertainty quality.  \textbf{Left:} 90\%-CI coverage by depth bin; the dashed line marks the nominal $90\%$ level.  GP~L0 maintains near-nominal coverage for shallow and medium cuts; GP~L2 degrades sharply at $0.3\text{--}1.0$\,mm.  \textbf{Right:} Mean 90\%-CI width.  The hybrid produces wider but less reliable intervals at medium and deep cuts.}
  \label{fig:depth_uq}
\end{figure}
\emph{Depth-dependent analysis.} Stratifying by milling depth (Fig.~\ref{fig:depth_uq}) reveals that these global figures average over distinct regimes. At shallow cuts ($|\bar d| < 0.3$\,mm) both models meet the $90\%$ target ($100\%$ and $89\%$). At intermediate depths ($0.3\text{--}1.0$\,mm) GP~L0 maintains ${\sim}88\%$ while GP~L2 drops to ${\sim}58\%$~--- the regime where physics residuals contain structured errors that the hybrid's variance does not capture. At deep cuts ($>1.0$\,mm) both degrade to ${\sim}67\%$. Notably, GP~L2 produces \emph{wider} intervals than GP~L0 at medium and deep cuts yet achieves \emph{worse} coverage, indicating that the deterministic physics model shifts the intervals rather than merely narrowing them. 
\emph{Physics contribution decomposition.} Each GP~L2 prediction decomposes as $\hat y = f_\text{phys}(x) + \hat r(x)$. The physics fraction varies from ${\sim}81\%$ at shallow cuts to ${\sim}60\%$ at deep cuts. This provides a transparent decision rule: when physics dominates, the prediction rests on calibrated scaling laws; when the residual is substantial, the GP predictive variance should be consulted. This interpretability~-- knowing \emph{how much} of a prediction is physics versus data-driven~--is the hybrid's most distinctive advantage, even when point accuracy is comparable.

\section{Discussion}
\label{sec:discussion}

The numerical results
are specific to AWJM on Inconel\,718. The methodological conclusions, however,
depend only on structural properties -- small~$n$, heterogeneous data quality, imperfect physics, and
evaluation sets too small for stable ranking -- shared by a broad class of precision manufacturing problems~\cite{Zhao2024}. Because all algorithms, curation hypotheses, and evaluation protocols are applied to the same dataset, each factor's effect can be isolated.

\subsection{Threats to Validity}

The study has four limitations. First, it covers one material, machine setup, and laboratory environment; external validity rests on problem structure, not multi-dataset replication. Second, 10-fold CV is more informative than a 15-point hold-out but does not eliminate small-sample variability; reported standard deviations are part of the result. Third, Stage-2 curation is hypothesis-driven, not neutral – the framework exposes that dependence rather than burying it in preprocessing. Fourth, the physics baseline requires material-specific calibration, so conclusions about physics integration apply at the level of integration \emph{strategy}, not as endorsement of one closed-form model. Replication on additional materials and processes remains the most important next step.

\section{Conclusion}
\label{sec:conclusion}
This paper studied how data curation and physics integration interact in
small-data regression for AWJM on Inconel 718 ($n=155$). A central methodological finding is that data preparation is not a neutral preprocessing step: separating physics-based cleaning from statistical curation reveals curation itself as a modelling hypothesis. For GP, the RMSE spread across hypotheses ($0.10$\,mm) exceeds the spread across algorithm classes ($0.07$\,mm at fixed curation); tree-based methods are less sensitive ($0.03$\,mm spread).
Physics integration is learner-dependent rather than uniformly beneficial. A compact physics backbone with GP residual learning yields competitive accuracy, lower fold-to-fold variance, and an
interpretable decomposition $\hat y(x)=f_{\mathrm{phys}}(x)+\hat r(x)$, whereas
construction degrades tree-based learners whose inductive bias is incompatible with the residual structure.  Standalone GP intervals are approximately calibrated ($86\%$ empirical coverage at nominal $90\%$), whereas the hybrid is systematically over-confident ($74\%$) because it omits physics model-form uncertainty.
Bayesian hyperparameter tuning improves parameter-sensitive baselines (GB, SVR) by up to $0.09$\,mm RMSE but degrades the hybrid GP pipeline, where ${\sim}16$ validation points per fold cannot guide a two-stage search reliably. Single-split rankings at this sample size are too fragile for strong conclusions. 
Overall, credible model comparison on small process datasets requires explicit curation hypotheses, resampling-based evaluation, and physics integration strategies matched to the learner's inductive bias. Future work should extend the hybrid approach to probabilistic physics baselines and test transfer across materials.

\paragraph{Acknowledgements.}
This work was funded by the Ministry of Economic Affairs of the State of North Rhine-Westphalia (Germany) under grant no. EFRE-20800516.
\bibliographystyle{splncs04}
\bibliography{references}

\end{document}